\documentclass[letterpaper, 10 pt, journal, twoside]{IEEEtran}                                    
\usepackage{amsmath, amsfonts} 
\usepackage{graphics}
    \DeclareGraphicsExtensions{.jpg,.pdf,.mps,.png} 
    \graphicspath{{fig/} {./}} 
\usepackage{epsfig} 
\usepackage{amssymb}  
\usepackage{bm} 
\usepackage[flushmargin,hang]{footmisc}
 
\usepackage{flushend}

\newcommand{\fig}[1]{Fig.~\ref{#1}}

\newcommand{\subfig}[2]{Fig.~\ref{#1}(\subref{#2})} 

\def\epsgaiji#1{\leavevmode\kern-0.025zw\raise-.37zh\hbox{%
  \epsfile{file=#1,width=1.05zw}}\kern-0.025zw}
\newcommand{\MARU}[1]{{\ooalign{\hfil#1\/\hfil\crcr\raise.167ex\hbox{\mathhexbox20D}}}}

\usepackage{array}

\usepackage{siunitx}
\usepackage{gensymb} 
\usepackage{multirow} 
\usepackage{caption}
\usepackage{subcaption}
\usepackage{pgfplots}
\pgfplotsset{compat=newest}
\usetikzlibrary{plotmarks}
\usetikzlibrary{arrows.meta}
\usepgfplotslibrary{patchplots}
\usepackage{xcolor}
\usepackage{grffile}
\pgfplotsset{plot coordinates/math parser=false}
\newlength\fwidth
\newlength\fheight

\usepackage{cite}

\begin{document}

\title{Enabling Faster Locomotion of Planetary Rovers \\with a Mechanically-Hybrid Suspension}

\author{David Rodríguez-Martínez$^{1*}$, Kentaro Uno$^{2*}$, Kenta Sawa$^{2}$, Masahiro Uda$^{2}$, Gen Kudo$^{2}$, Gustavo Hernan Diaz$^{2}$,\\ Ayumi Umemura$^{2}$, Shreya Santra$^{2}$, and Kazuya Yoshida$^{2}$
\thanks{Manuscript received: July 10, 2023; Revised October 12, 2023; Accepted November 5, 2023}%
\thanks{This paper was recommended for publication by Editor Clement Gosselin upon evaluation of the Associate Editor and Reviewers' comments. 
This work was supported by the Japanese Ministry of Education, Culture, Sports, Science \& Technology and the European Space Agency.}%
\thanks{$^{1}$D. Rodríguez-Martínez is with the Advanced Quantum Architecture Laboratory (AQUA) of the École Polytechnique Fédérale de Lausanne (EPFL), Switzerland 
        {\tt\footnotesize david.rodriguez@epfl.ch}}%
\thanks{$^{2}$K. Uno, K. Sawa, M. Uda, G. Kudo, G. H. Diaz, A. Umemura, S. Santra, and K. Yoshida are with the Space Robotics Lab. (SRL) in the Department of Aerospace Engineering, Graduate School of Engineering, Tohoku University, Japan
        {\tt\footnotesize unoken@tohoku.ac.jp}}%
\thanks{\textit{*David Rodríguez-Martínez and
Kentaro Uno contributed equally to this work. Corresponding author is David Rodríguez-Martínez.}}%
\thanks{Digital Object Identifier (DOI): 10.1109/LRA.2023.3335769.\\}

\thanks{\copyright2023 IEEE. Personal use of this material is permitted. Permission from IEEE must be obtained for all other uses, in any current or future media, including reprinting/republishing this material for advertising or promotional purposes, creating new collective works, for resale or redistribution to servers or lists, or reuse of any copyrighted component of this work in other works.}}

\markboth{IEEE Robotics and Automation Letters. Preprint Version. Accepted November, 2023}
{Rodríguez-Martínez \MakeLowercase{\textit{et al.}}: Enabling Faster Locomotion of Planetary Rovers with a Mechanically-Hybrid Suspension} 

\maketitle

\begin{abstract}
The exploration of the lunar poles and the collection of samples from the martian surface are characterized by shorter time windows demanding increased autonomy and speeds. Autonomous mobile robots must intrinsically cope with a wider range of disturbances. Faster off-road navigation has been explored for terrestrial applications but the combined effects of increased speeds and reduced gravity fields are yet to be fully studied. In this paper, we design and demonstrate a novel fully passive suspension design for wheeled planetary robots, which couples for the first time a high-range passive rocker with elastic in-wheel coil-over shock absorbers. The design was initially conceived and verified in a reduced-gravity (1.625\,m/s${^2}$) simulated environment, where three different passive suspension configurations were evaluated against steep slopes and unexpected obstacles, and later prototyped and validated in a series of field tests. The proposed mechanically-hybrid suspension proves to mitigate more effectively the negative effects (high-frequency/high-amplitude vibrations and impact loads) of faster locomotion ($\sim$1\,m/s) over unstructured terrains under varied gravity fields. 
\end{abstract}

\begin{IEEEkeywords}
Space Robotics and Automation; Compliant Joints and Mechanisms; Mechanism Design
\end{IEEEkeywords}

\IEEEpeerreviewmaketitle

\section{INTRODUCTION}
\label{sec:intro}

\begin{figure}[!h]
\centering
\includegraphics[width=\linewidth]{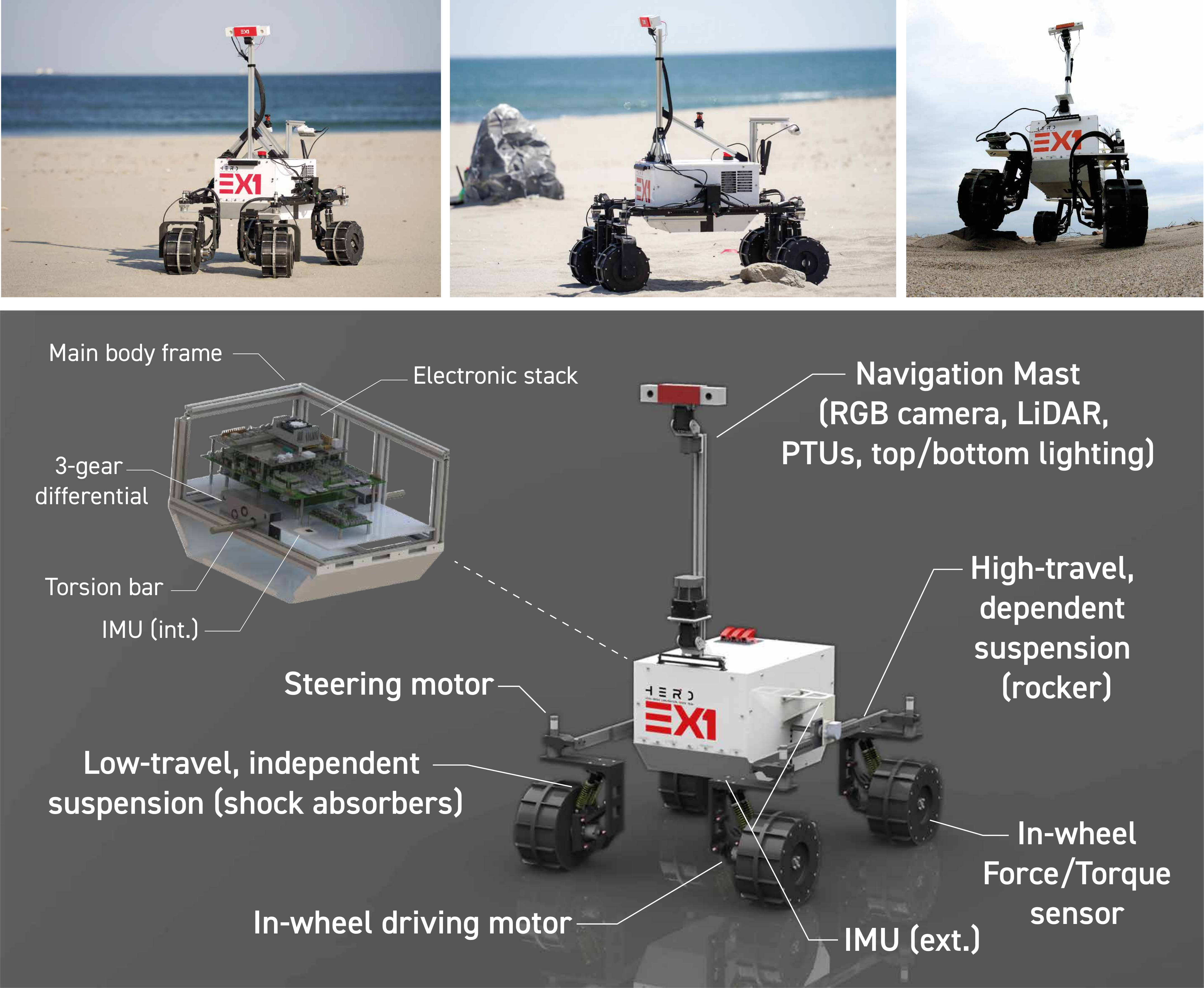}  
\caption{Our fast-moving planetary rover testbed, Explorer~1 (EX1). It has been designed with a mechanically-hybrid suspension consisting of a high-range dependent rocker suspension combined with low-range independent in-wheel coil-over shock absorbers, which enables smooth and stable driving at increased speeds ($\sim$1\,m/s) over loose soils, irregular hard surfaces, and hazard-abundant landscapes.}
\label{fig:ex1}
\end{figure}

\IEEEPARstart{R}{obots} have eased many of the tasks performed in space. They have assisted humans in building habitable labs in low-Earth orbit and have traversed the deserted lands of Mars in the name of science. A number of upcoming exploration missions require, however, robots capable of performing in uncharted domains for which technological innovations are necessary and doing so under increasingly limited time widows.

The growing interest in exploring the lunar poles serves as a good example. The extraction and use of hydrogen-bearing elements and other compounds \cite{Colaprete2010} could prove essential for the long-term sustainable exploration of space. However, unlike equatorial regions previously visited, the poles of the Moon harbor severe terrain elevation changes, extreme day-night temperature fluctuations, regions with limited natural illumination, and a sun that at times barely rises above the horizon \cite{Potts2015}. These features demand faster, more efficient, and highly autonomous robotic platforms capable of coping with a broad spectrum of environmental constraints unfaced by previous missions.

\subsection{Contributions}
\label{subsec:contributions}
In this paper, we present the design and prototyping of a new passive suspension concept, named mechanically-hybrid suspension (MHS), capable of enabling planetary robots to safely negotiate unstructured terrains at speeds that approach 1\,m/s (see \fig{fig:ex1})---almost two orders of magnitude faster than contemporary rovers. Our proposed design can be implemented on any robot regardless of size and wheel number (4-, 6-, and even 8-wheel configuration) and without increasing power demands. The design arises from our need to understand the impact of both increasing speeds and reduced gravity fields on the locomotor performance of mobile robots. This work addresses the following questions:

\begin{enumerate}
\item What is the impact endured by free-balancing suspensions when facing some of the salient features of the lunar surface at increasing speeds up to 1\,m/s?
\item What degree of improvement could be obtained from the addition of passive energy-dissipation devices?
\item How much do dependency and elasticity affect performance, and which configuration yields optimal results? 
\end{enumerate}

\subsection{Background}
\label{subsec:background}
Passive, inelastic, free-balancing suspensions in 4-to-8-wheeled chassis configurations have been employed by most of the rovers commissioned to explore the Moon and Mars. Suspensions used in space are optimized for supporting and evenly distributing the weight of the rover, allowing it to overcome irregular terrains and obstacles, and mitigating the effects of impacts and vibrations while isolating the sensitive optics and electronics onboard from these unwanted effects. Additionally, these suspensions are heavily constrained in terms of mass, volume, and power, making the rocker-bogie (RB) suspension \cite{Bickler1989} the most widely used type of suspension design. 

\begin{figure}[!h]
\centering
\includegraphics[width=0.8\linewidth]{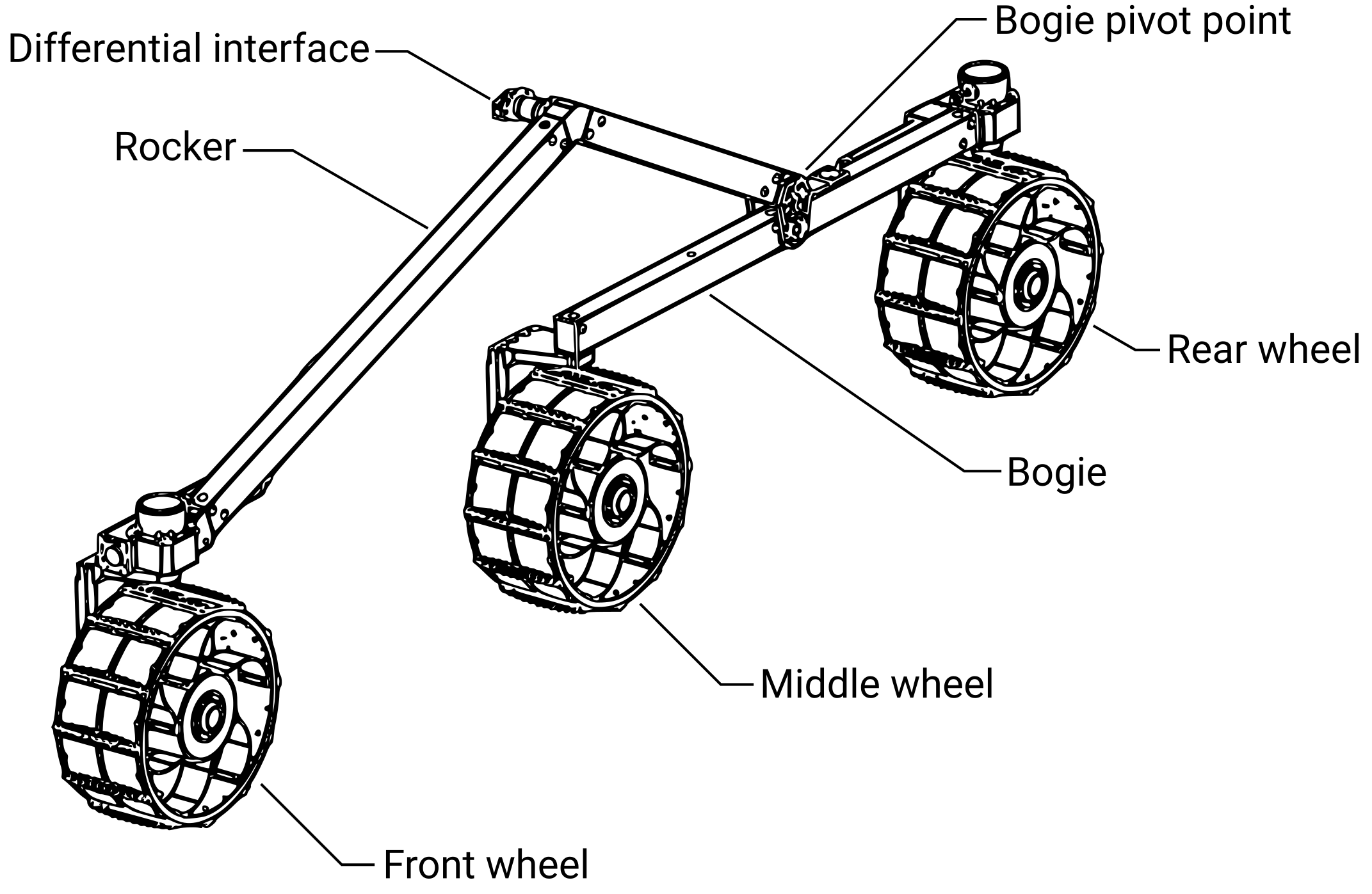}  
\caption{The rocker-bogie suspension.}
\label{fig:rocker-bogie}
\end{figure}

First developed in the frame of NASA's Mars Pathfinder mission \cite{Eisen1997}, the RB suspension consists of a mechanism of two linkages (see Fig.\,\ref{fig:rocker-bogie}). In the most commonly used configuration, a larger forward linkage called the rocker is fixed to the front wheel at one end and attached to the smaller rearward linkage, the bogie, at the other end through a free-rotating pivot point. Intended for 6-wheel configurations, the middle and rear wheels are each linked to both ends of the bogie. The rockers of both sides are connected together and attached to the chassis through a differential that maintains the body of the rover at a pitch angle equal to the average rotation of the two rockers.  

The RB suspension enables the rover to passively overcome irregular topographies and obstacles of a size comparable to its wheel diameter without losing contact with the ground. Due to the unconsolidated nature of the soils covering the vast majority of the lunar and martian surfaces, maintaining good traction levels and reducing slippage has been historically one of the biggest hurdles for planetary rovers \cite{Gonzalez2018}. NASA's rovers \textit{Sojourner}, which presented a reversed RB configuration (i.e., bogie facing forward), \textit{Spirit}, \textit{Opportunity}, \textit{Curiosity}, and \textit{Perseverance}, and China National Space Administration's (CNSA) \textit{Yutu-2} and \textit{Zhurong} rovers were all designed with an RB suspension. At the same time, these missions have been characterized by one significant limitation: speed.

The RB suspension was designed for operational speeds below $\sim$10 cm/s. At higher speeds, the structural integrity of the suspension and the stability of the robot cannot be ensured. Attempts have been made to broaden the range of applications of RB suspensions by independently controlling the speed of the wheels \cite{Miller2002} or dynamically adapting the suspension configuration \cite{Wang2016}. In an effort to find alternative solutions that are better suited to a wider range of environmental conditions and speeds, actively articulated and adaptive suspension designs have been widely discussed and proposed \cite{Iagnemma2003,Bartlett2008,Cordes2014}. While most of these solutions could be perfectly employed to maximize traversability and to minimize the detrimental effects resulting from high-speed locomotion, most rely on the optimal performance of other systems (e.g., hazard detection and terrain segmentation) or require an additional, non-negligible supply of power (e.g., to operate additional electromechanical actuators). 

Faster lunar vehicles are, however, not completely new. The Russian \textit{lunokhods} and NASA's two-crew-piloted Lunar Roving Vehicle (LRV) were capable of traveling at speeds of 0.5 and 5\,m/s \cite{Malenkov2016a, NASA1973a}, respectively. The \textit{lunokhods}, used during Luna 17 and Luna 21 missions, were designed with an 8-wheel suspension consisting of four carriers fixed to the bottom of the chassis, each hosting a pair of rigid wheels attached to each carrier via swing arms (allowing the wheel to move vertically while connected to the chassis) \cite{Malenkov2016a}. On the other side, the LRV implemented an independent double-wishbone suspension, frequently used in conventional road vehicles. Elasticity was provided through transverse torsion bars in both upper and lower control arms, in addition to compliant wheel rims; and damping was provided through a conventional silicone-oil damper \cite{NASA1971a}. Mission reports state the tendency of the suspension to bounce uncontrollably when traveling over surfaces with a large density of small craters (\textless\,1\,m) and the impossibility of steering effectively, i.e., without excessive side slip, at speeds above 1.4\,m/s \cite{NASA1973a}. In both cases, the barren and subdued lunar landscape and the need to drive at times directly toward the Sun did not make negotiating these obstacles any easier. The capability to drive faster of both platforms seemed to be closely associated with their ample power reserves and the direct human input---both vehicles were either directly piloted or teleoperated from Earth---rather than with variations purposely introduced in their suspension designs \cite{RodriguezMartinez2019}. 

\section{PRELIMINARY ANALYSIS}
\label{sec:analysis}
During the Apollo and Luna missions, data on the suspension performance were never collected. Subjective evaluations on rideability at higher speeds are insufficient to argue in favor of one or another suspension configuration. This is particularly relevant in the case of autonomous robots where the level of impact loads and vibrations gain significance while aspects such as comfort can be neglected. We needed to first understand quantitatively the effects that increasing speeds---up to 1 m/s---would have on different suspension configurations and the potential improvements or limitations associated with the addition of passive energy dissipation devices under reduced-gravity, unstructured environments. To this end, we developed a series of simulation modules running on Coppelia Robotics' simulator CoppeliaSim \cite{vrep2013} in combination with CM Labs' high-fidelity physics engine Vortex.

\subsection{Multibody dynamic model}
\label{subsec:mbd_model}
\begin{figure}[!t]
\centering
\includegraphics[width=\linewidth]{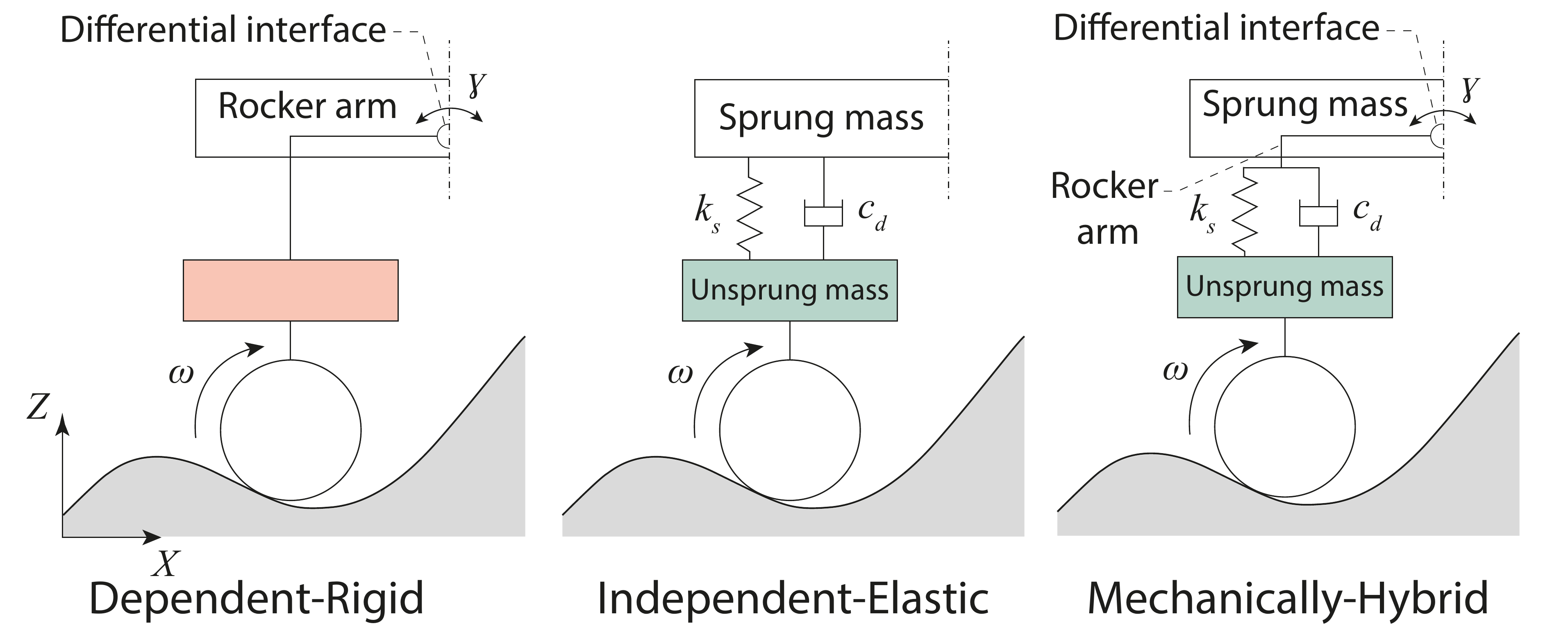}  
\caption{Simplified quarter-car system diagrams of the three different suspension configurations used during the preliminary analysis. Design properties were considered constant throughout the analysis and included stiffness, $k_s$, damping, $c_d$, and the length and rotation angle of the rocker, $\gamma$, which are in turn related to its travel range.}
\label{fig:suspensions}
\end{figure}

To establish the foundation for our comparative analysis, we defined three dynamic models of a 4-rigid-wheel rover featuring All-Wheel Drive/2-Wheel Steering (AWD/2WS), each incorporating a distinct passive suspension configuration (see \fig{fig:suspensions}):
 
\begin{itemize}
    \item A Dependent-Rigid (DR) model was used to benchmark our analysis. It makes use of a passive rocker on each side linked by a differential;
    \item An Independent-Elastic (IE) model implements a shock absorber at each wheel simulating the suspension of conventional terrestrial vehicles;
    \item And, last, an unprecedented conceptual model we called Mechanically-Hybrid Suspension (MHS). 
\end{itemize}

The concept of the MHS stems from the hypothesis that combining the high travel range and stable load distribution of dependent rockers with the capacity to attenuate impacts and vibrations of shock absorbers would improve the traversability and reliability of faster-moving planetary robots. No previous planetary rover and, to our best knowledge, no conceptual study has ever explored the possibility of combining in one single design these two design principles.

\begin{table}[b]
\centering
\caption{Dynamic model properties.}
\label{table:rover_model}
\begin{tabular}{p{3cm}p{2cm}} 
\hline
Property & Value \\
\hline
Wheel radius & 0.1 m \\ 
Wheel-base (arm length) & 0.6 m (0.3 m)\\ 
Wheel-track & 0.47 m \\  
Height of CoM. & 0.25 m \\
Spring ratio & 2 kN/m \\
Damping coefficient & 350 Ns/m \\
Spring free-length & 35 mm \\
Wheel load (front/rear) & 4.8/5.0 kg \\
Total mass & 19.6 kg\\
\hline
\end{tabular}
\end{table}
We opted for a 4-wheel configuration as it represents the simplest and most widely-used locomotor configuration in autonomous ground vehicles. Nonetheless, the results of this preliminary analysis can be extrapolated to 6- and 8-wheel configurations. Each suspension configuration presented the same mass distribution and suspension kinematics, both fixed throughout the comparative study and based on the design specifications of ElDorado-2, a long-standing robotic platform developed at Tohoku University's Space Robotics Lab (see Fig.~\ref{fig:modules}). General characteristics of the rover model are presented in Table~\ref{table:rover_model}. Spring-damper systems in CoppeliaSim were simulated by means of prismatic joints whose reaction force is controlled by a proportional-derivative controller in which the proportional and derivative gains are replaced by the spring ratio, $k_s$, and damping coefficient, $c_d$, respectively. The specific values used to describe the elastic part of the suspension resulted from a sensitivity analysis aimed at acquiring a stable movement and a limited static deflection of the rover model. Note that while optimizing stiffness and damping is vital in the ultimate design of an elastic suspension system, this would entail analyzing the response of the rover to a specific set of excitations based on mission-driven and design-specific requirements. This analysis would make the outcome non-generalizable and, therefore, it has been considered out of the scope of our study. 

\subsection{Simulation modules}
\label{subsec:modules}
\begin{figure}[!h]
\centering
\includegraphics[width=\linewidth]{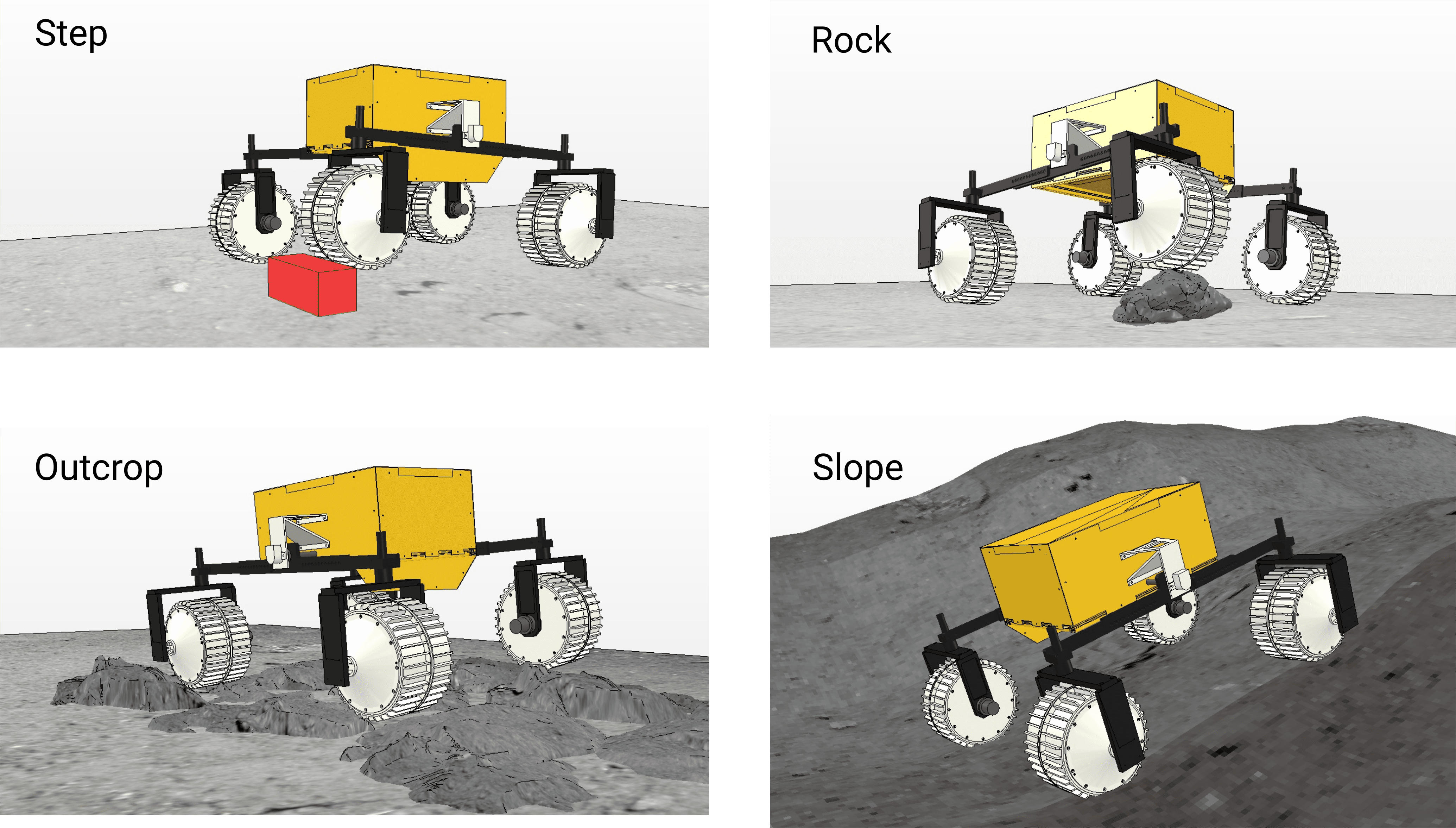}  
\caption{Visual layer of the four multibody dynamics simulation modules including a model of the ElDorado-2 rover used in this study.}
\label{fig:modules}
\end{figure}

We developed a series of obstacle negotiation and gradeability modules. The environmental characteristics of each module were based on characteristic features of the surface of the Moon (see Fig.\,\ref{fig:modules}). A gravity field of 1.625\,$\mathrm{m/s^2}$ representative of the lunar surface was used for all our simulations. To focus exclusively on the performance of the suspension, no closed-loop traction or motion control was implemented. 

The obstacle negotiation module consisted of step obstacles of increasing height (1--12\,cm), a dynamically-enabled 10-cm hemispherical rock, and a 1.5-m long outcrop---i.e., a partially exposed section of bedrock---with protrusions as high as 10\,cm, all modeled within CoppeliaSim.

The gradeability module presented 1.5-m slopes of increasing inclination from 5$^\circ$ up to a maximum of 30$^\circ$. For the sake of maintaining the comparative analysis within reasonable margins, only situations where the robot faced the slopes at a heading angle of 0$^\circ$ (straight climbing) were simulated. 

\subsection{Wheel-soil contact model}
\label{subsec:wscontact}
The lunar surface is characterized by a top layer of fine-grained, slightly cohesive regolith. While the locomotor performance of an off-road vehicle is defined by the combination of the suspension and terramechanical performances---i.e., the capacity to gain and/or maintain traction---, in this study we focus exclusively on the former. For the sake of simplicity, and in order to have a symbolic representation of the frictional behavior of the lunar soil, we opted for a Coulomb friction model with an isotropic friction coefficient of 0.4 for the wheel-soil contact---frequently used as representative of metallic wheels rolling over sandy terrains---and 1.0 for the interaction with obstacles such as rocks and outcrops. 
 
\subsection{Performance evaluation parameters}
\label{subsec:param}
We based the comparative evaluation of the performance on the success of each configuration in overcoming a given obstacle, the maximum vertical load and pitch torque measured at both ends of the rocker arms, and the maximum vertical acceleration experienced by the chassis. 

\subsection{Results and discussion}
\label{subsec:results}
\subsubsection{Obstacle negotiation performance}
\label{subsubsec:onp}
\begin{figure}[!h]
\centering
\includegraphics[width=\linewidth]{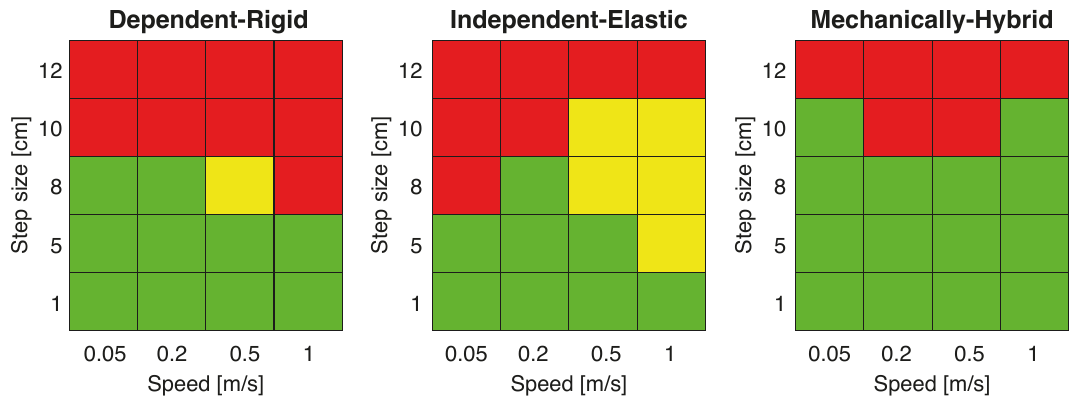}  
\caption{Series of heatmaps displaying the success (green), semi-success (yellow), or failure (red) of the different suspensions in overcoming steps of increasing height (rows) at different speeds (columns). Semi-success defines situations where only the front wheels overcame the step.}
\label{fig:step_hm}
\end{figure}

The heatmaps shown in Fig.~\ref{fig:step_hm} represent the level of success of the different suspension configurations in overcoming perfect steps of height 1--12\,cm(rows) at speeds ranging from 0.05--1\,m/s(columns). 

\begin{figure}[!h]
\centering
\includegraphics[width=\linewidth]{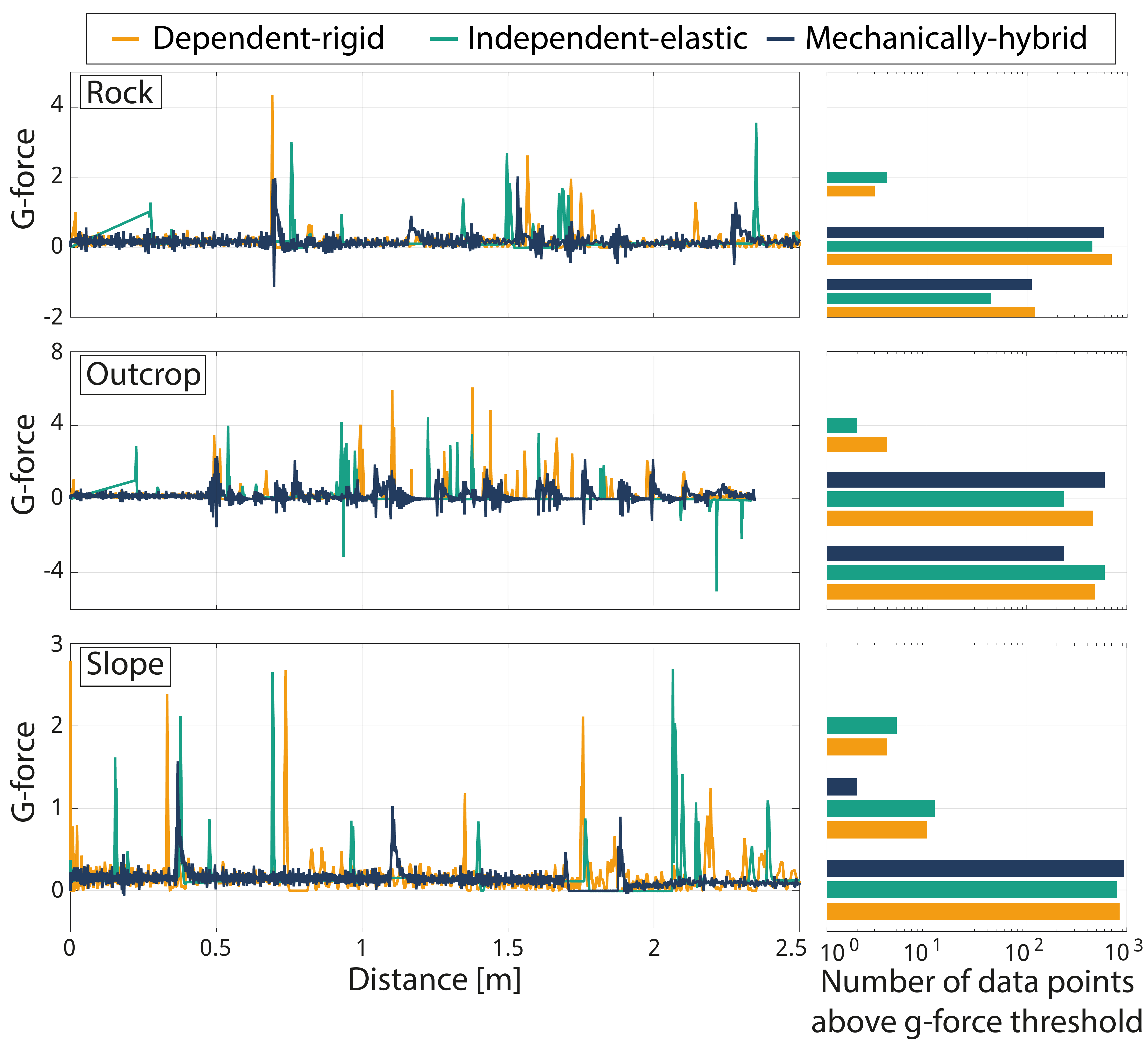}  
\caption{Vertical accelerations of the rover chassis recorded throughout the course of the interaction with multiple obstacles at 1\,m/s under lunar gravity (0.166\,g), namely: a 10-cm hemispherical rock (top), a 1.5-m outcrop (middle), and a 20$^\circ$ slope (bottom). The degree of attenuation of vibrations drastically increases with the use of the MHS.}
\label{fig:gforce}
\end{figure}

Due to their vertical profile, steps are considered the most challenging obstacle to negotiate. Counterintuitively, the results displayed in \fig{fig:step_hm} present an increase in the rate of success with speed when elasticity is incorporated into the design. This outcome can be associated to the effect of the suspension rebound, which exerts an upward force (whose value increases with speed due to increasing compression loads) aiding the rover tractive forces in overcoming the obstacle. This effect can be more clearly seen in the vertical acceleration profiles displayed in \fig{fig:gforce}. In this case, only the most extreme situations are illustrated (largest obstacles and steepest incline). Negative g-force values have been recorded due to the compression and subsequent expansion of the elastic part of the suspension in the IE and MHS configurations when overcoming obstacles. The negative effects of suspension rebound can be reduced by the addition of dependency, as seen in the MHS case, and could be further mitigated with a deeper optimization of the suspension parameters.

 Table \ref{table:max_val} lists the maximum values of vertical load, pitch torque, and acceleration experienced at top speed (1\,m/s) for every obstacle type and for each configuration. Overall, the results obtained from the obstacle negotiation module illustrate the considerable benefits gained by introducing energy dissipation devices in the design of passive suspensions. On average among all the cases analyzed (steps, rocks, and outcrops), a 71\% reduction in maximum impact load, a 37\% reduction in maximum pitch torque, and a 33\% reduction in maximum vertical acceleration of the chassis were observed when elasticity and damping were incorporated into the design. While these are well-known behaviors, when dependency via free-balancing rockers is now included alongside elasticity and damping, the MHS outperformed the other two in every situation, overall reducing by 62\% and 43\% the detrimental effects of an irregular terrain when compared to the DR and IE configurations, respectively. The compliance of the in-wheel suspension attenuates the high-frequency/high-amplitude vibrations while the dependency of the rocker provides a more efficient weight transfer allowing the rover to overcome large obstacles without inducing excessive traction losses or instabilities. 

\begin{table*}[!hbt]
\centering
\caption{Maximum values of vertical impact load ($F$), pitch torque ($T$), and vertical acceleration ($Acc$) of the chassis for each suspension configuration. All results are with a reduction ratio from the worst one to the best.}
\label{table:max_val}
\begin{tabular}{p{2cm}|p{0.8cm}p{0.8cm}p{0.8cm}p{0.8cm}|p{0.8cm}p{0.8cm}p{0.8cm}p{0.8cm}|p{0.8cm}p{0.8cm}p{0.8cm}p{0.8cm}} 
\hline
Module & $F_{DR}$ & $F_{IE}$ & $F_{MHS}$ & Rate & $T_{DR}$ & $T_{IE}$ & $T_{MHS}$ & Rate & $Acc_{DR}$ & $Acc_{IE}$ & $Acc_{MHS}$ & Rate\\
\hline
10-cm rock & 8633.0 & 549.0 & \textbf{341.1} & -96\% & 88.3 & 63.4 & \textbf{62.0} & -30\% & 4.36 & 3.56 & \textbf{2.02} & -54\% \\ 
1.5-m outcrop & 7076.5 & 917.0 & \textbf{337.7} & -95\% & 156.7 & 98.0 & \textbf{58.1} & -63\% & 6.07 & 5.04 & \textbf{2.34} & -61\% \\ 
20$^\circ$ slope & 377.2 & 409.9 & \textbf{132.4} & -65\% & 58.4 & 49.9 & \textbf{31.2} & -47\% & 2.79 & 2.69 & \textbf{1.57} & -44\% \\  
\hline
\end{tabular}\\
\vspace{0.5mm}
\tiny{$D\!R\equiv$dependent-rigid, $I\!E\equiv$independent-elastic, $M\!H\!S\equiv$mechanically-hybrid suspension. \textbf{Units}: $F$[N], $T$[Nm], $Acc$[g].}
\end{table*}

\subsubsection{Gradeability}
\label{subsubsec:grade}
\begin{figure}[!h]
\centering
\includegraphics[width=\linewidth]{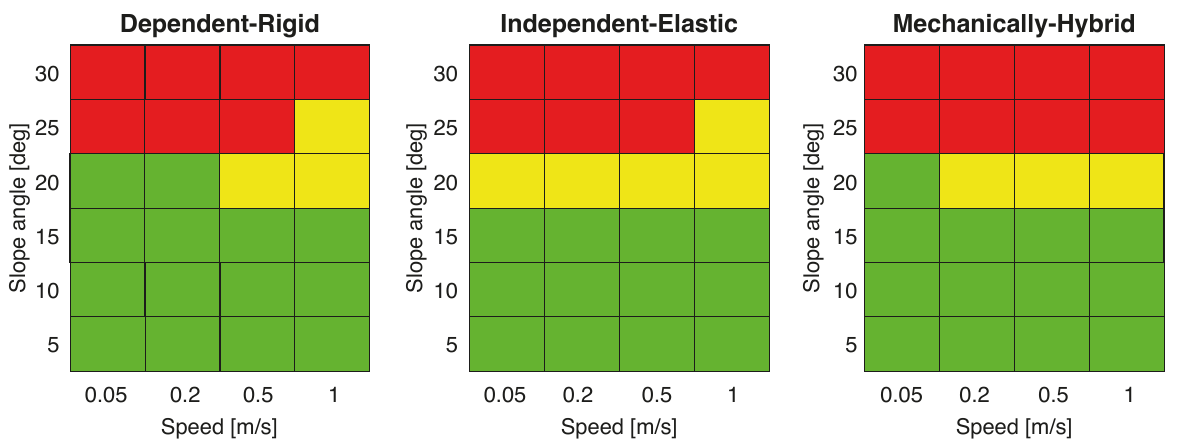}  
\caption{Series of heatmaps displaying the success (green), semi-success (yellow), or failure (red) of the different suspensions in overcoming slopes of increasing inclination (rows) at different speeds (columns). Semi-success defines situations where only the rear wheels maintained full contact with the ground throughout the climb.}
\label{fig:grade_hm}
\end{figure}

Less variation in the level of success of the different configurations was observed when the rover faced 1.5\,m slopes of varying inclination (5--30$^\circ$) at speeds ranging from 0.05--1\,m/s (see \fig{fig:grade_hm}). At higher speeds and regardless of the configuration, the top of the steepest slopes (20$^\circ$ and 25$^\circ$) were often reached with just the rear wheels in contact with the ground (semi-success cases \fig{fig:grade_hm})---a predominant effect in the IE and MHS configurations due to the excessive rebound of the suspension upon first confronting the slope.

We initially expected greater levels of variation in the ability to climb steeper inclines of the different configurations given the evidence presented in \cite{Gonzalez2019a}. In this work, the climbing ability of rocker arms evidently outperformed that of independent swing arms under every circumstance evaluated. Due to the slip-dominant nature of the vehicle-ground interaction when climbing a slope, it is possible that the absence of a more accurate representation of the wheel-soil contact behavior in our simulation modules and the lack of an active control scheme resulted in the lack of variation in the levels of success of the different suspension configurations. Despite this similarity, and in line with the observations previously made, the use of the MHS configuration managed to attenuate impact loads and vibrations beyond what was accomplished by either the DR or IE configurations. As an example, in the case of the rover facing a 20$^\circ$-slope at 1\,m/s (see Table \ref{table:max_val}), impact loads, pitch torque, and vertical accelerations of the chassis were reduced by 66.5\%, 42\%, and 43\%, respectively and on average, compared to the DR and IE configurations.

\section{SYSTEM DESIGN}
\label{sec:design}
In light of the evidence provided by the results of the simulations, we conceived a new rover prototype, dubbed Explorer 1 (EX1), based on the principles of the MHS configuration (see Fig. \ref{fig:ex1}). 

The MHS was designed to reliably cope with a higher degree of perturbations while maintaining the mechanical simplicity and reliability of the locomotion system and without excessively increasing the rover mass or its power requirements. EX1 presents a 4-wheel AWD/4WS locomotion configuration capable of achieving a maximum operational speed of 1\,m/s. Aluminum rocker arms are linked together and attached to the chassis through a 3-gear differential box housed inside the body frame. These rocker arms have a range of motion of about $\pm$250\,mm (2.5 times its wheel radius), only limited by the length of the wire harness of the actuator drive electronics. Attached at both ends of each rocker is a double-coil-over elastic suspension providing a lower travel range of 35\,mm. The harmonic drives of the steering motors act as the connecting pieces between the rocker arms and the compliant component of the suspension. This allows the latter part of the suspension to rotate with the wheel during steering, but has the inconvenience of varying the scrub radius---the distance between the steering axis and the vertical centerline of the wheel---based on the level of compression of the suspension; a shortcoming that was assumed in favor of the modularity of the design (i.e., the design is easily adaptable to 6- and 8-wheel configurations) and due to the short free-length of the damper. 

\begin{figure}[!h]
\centering
\includegraphics[width=\linewidth]{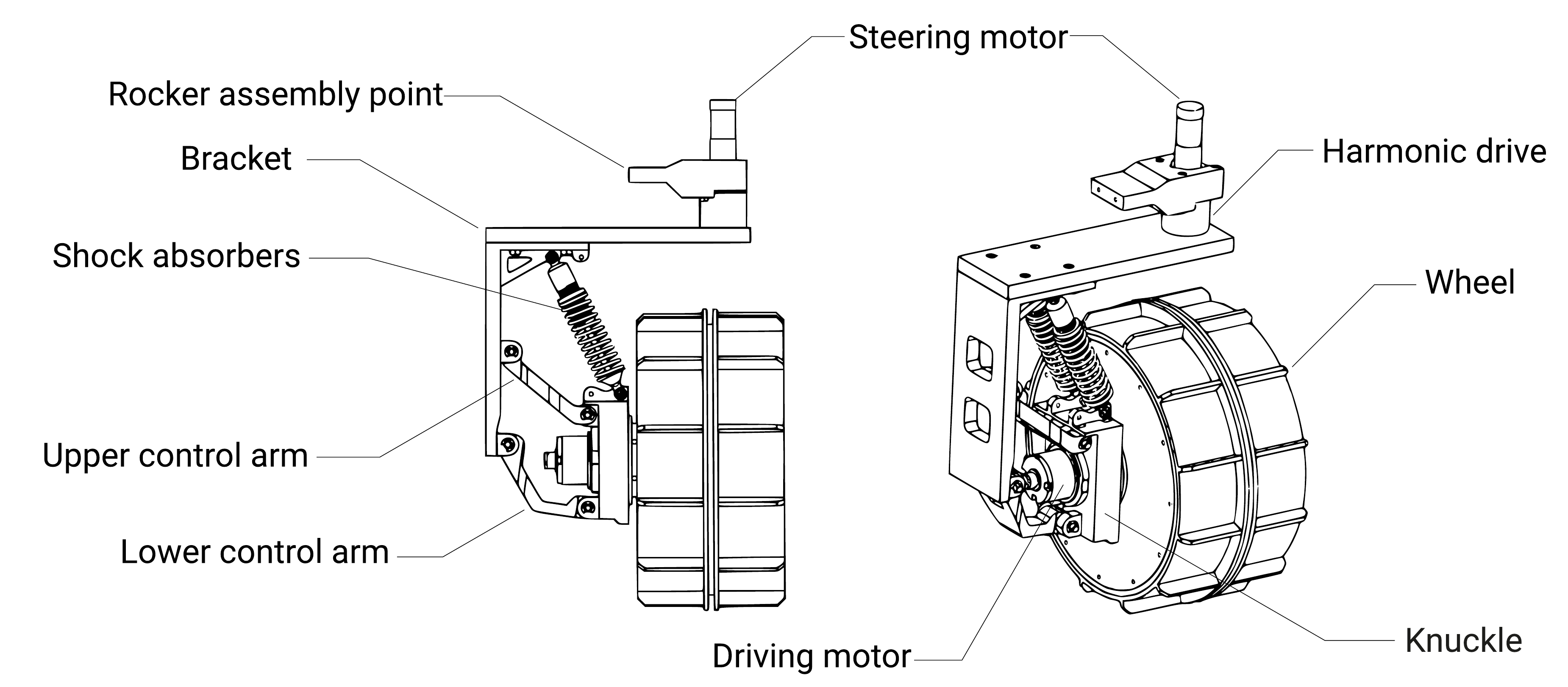}  
\caption{Design of the low-range elastic part of EX1 mechanically-hybrid suspension.}
\label{fig:lts}
\end{figure}

\begin{figure}[!h]
\centering
\includegraphics[width=0.9\linewidth]{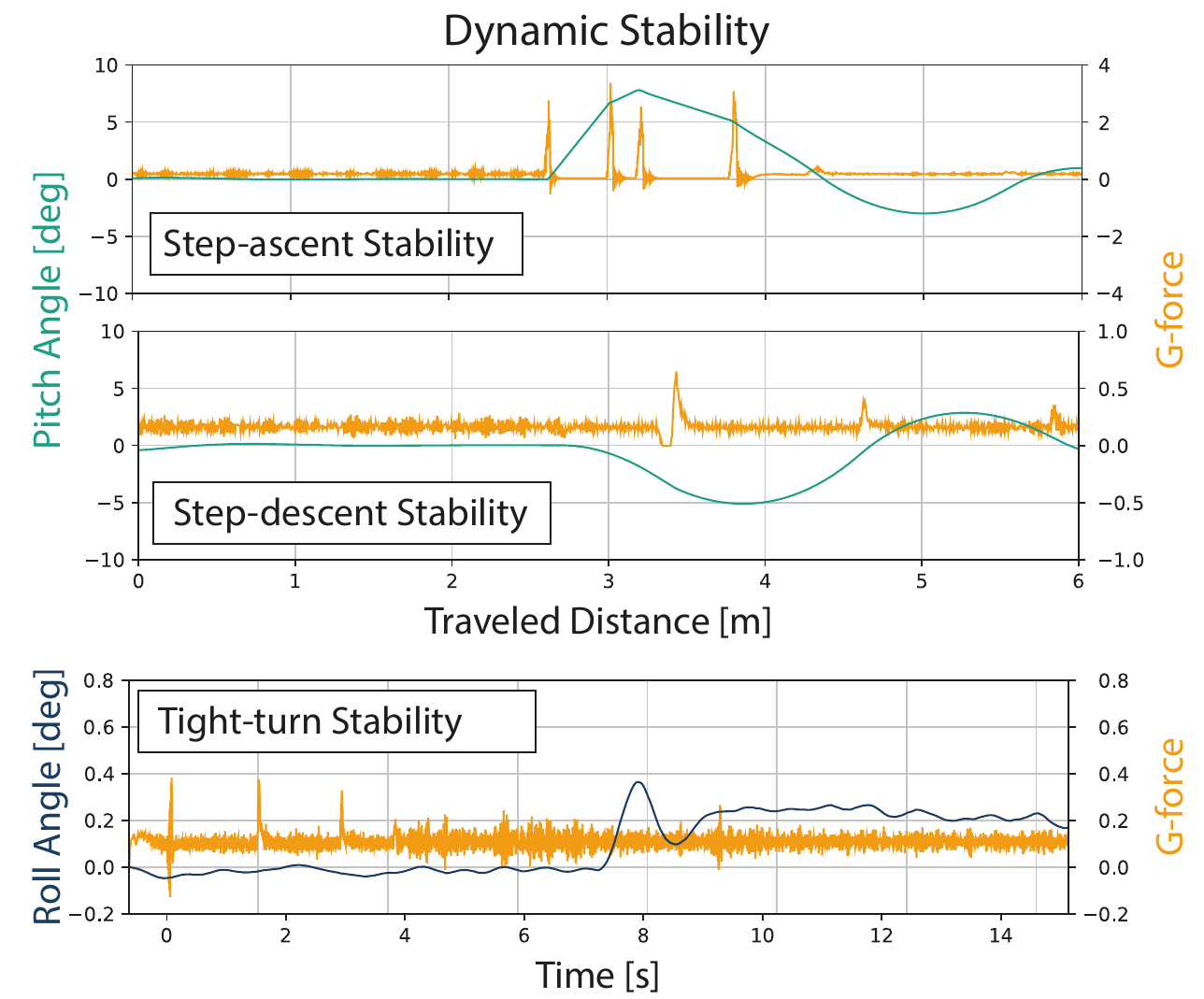}  
\caption{ Results of the dynamical response of the rover when negotiating a 5-cm step and a 3-m 90$^\circ$ left turn both performed at 1\,m/s.}
\label{fig:stability}
\end{figure}

The low-range (i.e., max. vertical travel in the order of \textit{mm}) suspension consists of upper and lower control arms passively commanded by a pair of 104-mm shock absorbers connected to the top of the wheel knuckle (see \fig{fig:lts}). This arrangement maintains the camber angle nearly constant during wheel travel. Two parallel shock absorbers were used to reduce the stiffness required on the springs while providing a certain level of redundancy in the design. The shock absorbers are formed by a replaceable 2.5\,kN/m spring (5\,kN/m per wheel) and an adjustable damper and were selected off-the-shelf from a radio-control car manufacturer. Both the bracket and the wheel knuckle were designed with multiple mounting points so that the overall stiffness of the suspension can be slightly modified by tilting the orientation of the dampers. The stability limits of the design were verified in simulations, achieving a static longitudinal/lateral stability under lunar gravity of 30$^\circ$ and a quick and smooth response to dynamic perturbations such as steps and cornering maneuvers (see \fig{fig:stability}).  

\section{FIELD TEST RESULTS}
\label{sec:results}
We conducted a field test campaign in a representative sandy field located in a nearby seashore to validate the locomotor performance of EX1. While these tests allow us to functionally validate a new suspension design for ground testing purposes, it should be noted from the outset the inadequacy of our approach to testing for the optimization of design parameters with respect to a potential flight model configuration. The conventional approach to validating the mobility performance of planetary rovers on Earth prior to their missions suffers from a strong limitation in situations when speed plays a role. While gravity scaling is often applied to testing platforms---i.e., adapting the mass of engineering models to represent the overall weight of the flight model at destination---observable behaviors under testing are only representative when the quasi-static approximation can be applied. The moment dynamic effects dominate the behavior of the rover and its interaction with the ground \cite{RodriguezMartinez2019a}, as is the case with our experiments, the full-body mass of the rover shall be used for a representative characterization of the performance and subsequent optimization of design parameters. This drastically affects the rover response to environmental and operational stimuli \cite{Biesiadecki2006a}. In these cases, gravity offloading must be applied \cite{orr2022effects} but further work would still be required to properly model the complex interplay of inertial, gravitational, and frictional forces taking place.

\subsection{Dynamic stability}
\begin{figure}[!th]
\centering
\includegraphics[width=\linewidth]{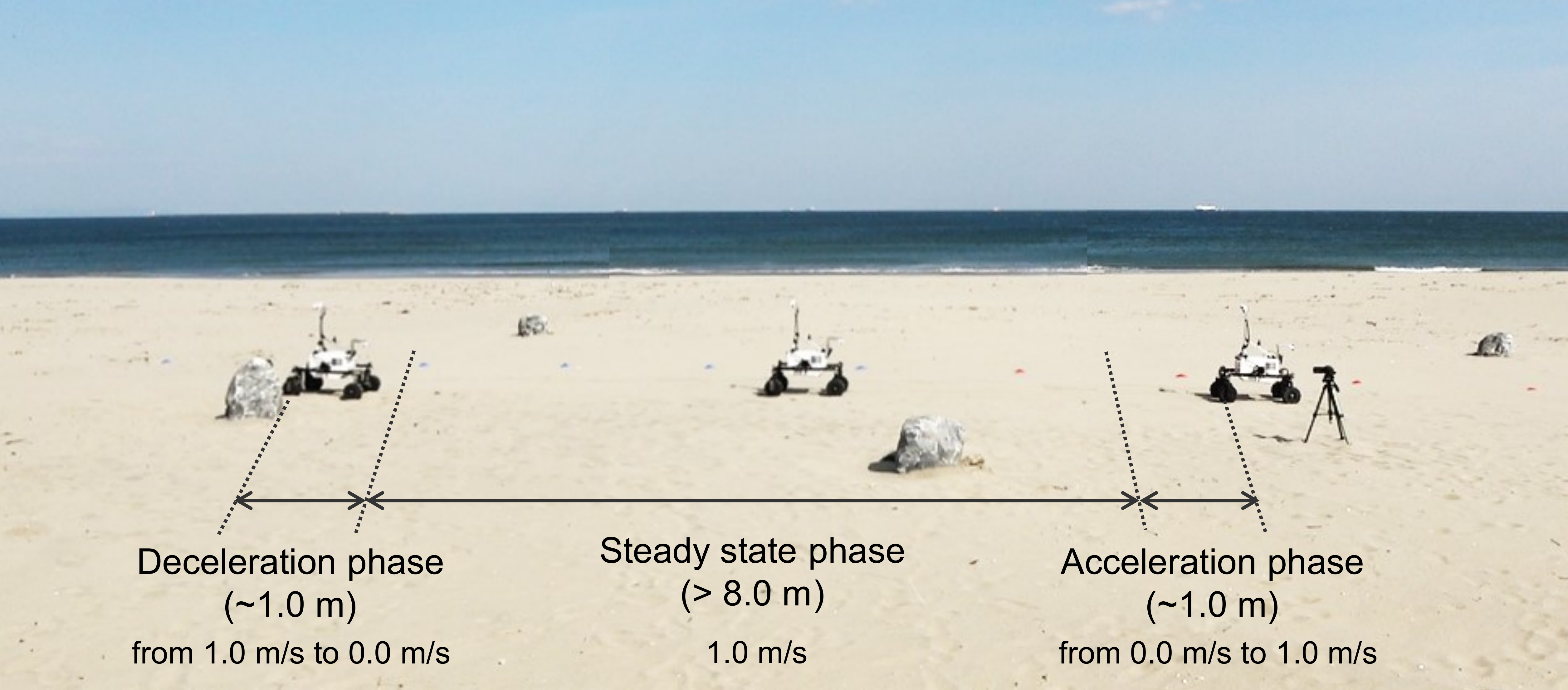}  
\caption{Dynamic stability test setup and phases.}
\label{fig:maneuverability_field_test}
\end{figure}

The first experiment was aimed at evaluating the contribution of the independent shock absorbers when moving at high speed over a 10-m, nearly-flat, unconsolidated, and dry sand field in both transient and steady-state conditions. In this case, the rover was commanded to follow a straight trajectory divided into three phases: a) an acceleration phase where the rover speeds up to 1.0\,m/s, b) a second steady-state phase at a uniform speed of 1.0\,m/s, and c) a final phase where the rover is decelerating to a full stop (see \fig{fig:maneuverability_field_test}).

We performed these tests with the rover in three different suspension configurations representative of the DR, IE, and MHS configurations used in the simulations. In the DR configuration, the rocker of EX1 is free to rotate but the shock absorbers are replaced by rigid elements locking the low-range suspension in place; while in the IE configuration, the free rotation of the rocker is now locked in its horizontal position while each wheel presents the elastic suspension described in \fig{fig:lts}. It is important to note that by locking the free-rotation of the rockers, each arm becomes a potential torsion bar in the IE configuration contributing to the dissipation of loads. However, this effect is negligible due to the overall size of the arms and distance from the wheel axis. Six runs were conducted for each suspension configuration. The ground was raked and leveled before every test run. All runs were performed on the same day under minor variations in temperature and humidity. Table~\ref{table:test_acc} lists the result of comparing the three configurations based on the vertical acceleration of the chassis as recorded by an IMU fixed to the top of the attachment element of the left-side rocker (see \fig{fig:ex1}). Max. and min. values, and the mean of the standard deviation of the vertical accelerations were computed from the raw data across all six runs.

\begin{table}[!hbt]
\centering
\caption{Comparison of absolute max., max.$-$min. gap, and mean of the standard deviation $\bar{\sigma}$ of the vertical accelerations across all runs experienced by the chassis for each configuration. The bold font means the best value.}
\label{table:test_acc}
\begin{tabular}{c c l r r r r} 
\hline
Test & Phase & Metric [g] & \textit{DR} & \textit{IE} & \textit{MHS} & Rate\\
\hline
\multicolumn{1}{c|}{\multirow{9}{*}{\begin{tabular}[c]{@{}c@{}}\textit{Dynamic} \\ \textit{Stability}\end{tabular}}} & \multicolumn{1}{l|}{\multirow{3}{*}{\textit{Accel.}}} & Max. & 1.81 & 1.47 & \textbf{1.44} & $-$20\% \\ 
 \multicolumn{1}{l|}{} & \multicolumn{1}{l|}{} & Max.$-$Min. & 1.55 & 1.00 & \textbf{0.98} & $-$37\% \\ 
 \multicolumn{1}{l|}{} & \multicolumn{1}{l|}{} & $\bar{\sigma}$ & 0.31 & 0.19 & \textbf{0.18} & $-$43\% \\  
\cline{2-7}
 \multicolumn{1}{l|}{}  & \multicolumn{1}{l|}{\multirow{3}{*}{\begin{tabular}[c]{@{}c@{}}\textit{Steady-} \\ \textit{state}\end{tabular}}} & Max. & 1.56 & 1.51 & \textbf{1.37} & $-$12\% \\ 
 \multicolumn{1}{l|}{} & \multicolumn{1}{l|}{} & Max.$-$Min. & 1.17 & 1.14 & \textbf{0.84} & $-$28\% \\ 
 \multicolumn{1}{l|}{} & \multicolumn{1}{l|}{} & $\bar{\sigma}$ & 0.19 & 0.17 & \textbf{0.14} & $-$28\% \\  
\cline{2-7}
 \multicolumn{1}{l|}{}  & \multicolumn{1}{l|}{\multirow{3}{*}{\textit{Decel.}}} & Max. & 1.41 & 1.28 & \textbf{1.26} & $-$11\% \\ 
 \multicolumn{1}{l|}{}  & \multicolumn{1}{l|}{}& Max.$-$Min. & 0.83 & 0.59 & \textbf{0.54} & $-$35\% \\ 
 \multicolumn{1}{l|}{}  & \multicolumn{1}{l|}{}& $\bar{\sigma}$ & 0.17 & \textbf{0.11} & \text{0.12} & $-$31\% \\  
\hline
\multicolumn{1}{c|}{\multirow{9}{*}{\begin{tabular}[c]{@{}c@{}}\textit{Obstacle} \\ \textit{Negotiation}\end{tabular}}} & \multicolumn{1}{c|}{\multirow{3}{*}{\textit{0.2\,m/s}}} & Max. & n/a & 1.37 & \textbf{1.20} & $-$12\% \\ 
 \multicolumn{1}{l|}{} & \multicolumn{1}{l|}{} & Max.$-$Min. & n/a & 0.75 & \textbf{0.39} & $-$48\% \\ 
 \multicolumn{1}{l|}{} & \multicolumn{1}{l|}{} & $\bar{\sigma}$ & n/a & 0.09 & \textbf{0.05} & $-$44\% \\  
\cline{2-7}
 \multicolumn{1}{l|}{}  & \multicolumn{1}{c|}{\multirow{3}{*}{\textit{0.5\,m/s}}} & Max. & n/a & 1.42 & \textbf{1.31} & $-$8\% \\ 
 \multicolumn{1}{l|}{} & \multicolumn{1}{l|}{} & Max.$-$Min. & n/a & 0.93 & \textbf{0.66} & $-$29\% \\ 
 \multicolumn{1}{l|}{} & \multicolumn{1}{l|}{} & $\bar{\sigma}$ & n/a & 0.20 & \textbf{0.19} & $-$5\% \\  
\cline{2-7}
 \multicolumn{1}{l|}{}  & \multicolumn{1}{c|}{\multirow{3}{*}{\textit{1.0\,m/s}}} & Max. & n/a & n/a & \textbf{1.47} & -\\ 
 \multicolumn{1}{l|}{}  & \multicolumn{1}{l|}{}& Max.$-$Min. & n/a & n/a & \textbf{0.98} & - \\ 
 \multicolumn{1}{l|}{}  & \multicolumn{1}{l|}{}& $\bar{\sigma}$ & n/a & n/a & \textbf{0.30} & - \\  
\hline
\end{tabular}\\
\vspace{0.5mm}
\tiny{$D\!R\equiv$dependent-rigid, $I\!E\equiv$independent-elastic, $M\!H\!S\equiv$mechanically-hybrid suspension.}
\end{table}

Results confirm an overall reduction of the vertical accelerations experienced by the chassis when the MHS is used compared to the other two configurations. This is particularly significant during the acceleration phase where the rover experienced greater vibrations due to an observed increase in wheel slippage. Our results are aligned with a well-established understanding of the performance of elastic suspensions in off-road terrestrial vehicles but it was important to evaluate the potential interference that in-wheel shock absorbers could have on the movement of free-balancing rockers, less common in terrestrial applications. Comparing the rover's ground truth trajectory as recorded by a total station and each wheel encoder confirmed that all three suspension configurations experienced similar slip ratio profiles, which increased up to 40\% during the acceleration phase and decreased to a maximum of 10\% during the steady-state phase. There is also no considerable difference between the three suspensions in lateral sliding: the rover always experienced around 10\% lateral sliding in the same direction, which could be due to the natural gradient of the terrain. These results were expected due to the minimum effect that the suspension design has on the mitigation of slippage compared to active traction control approaches.

\subsection{Obstacle negotiation capability}

\begin{figure*}[!hbt]
\centering
\includegraphics[width=\linewidth]{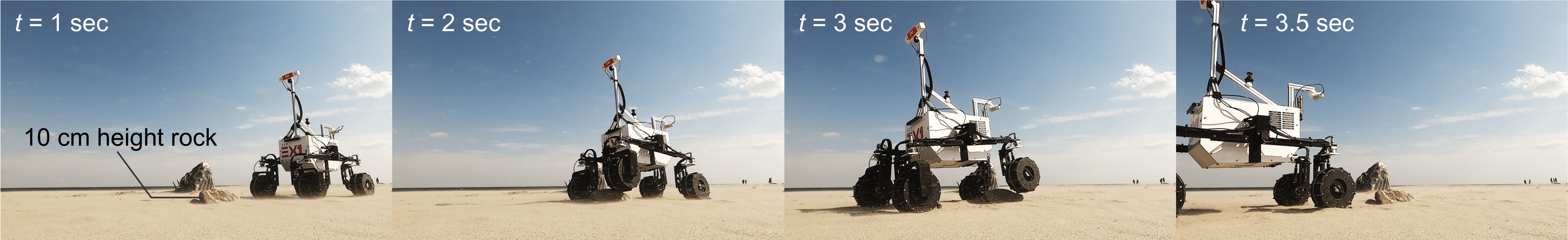}  
\caption{Representative images of EX1 overcoming a 10 cm rock at 1\,m/s as part of the obstacle negotiation tests (MHS).}
\label{fig:obstacle_negotiation_field_test}
\end{figure*}

In this second experiment, the rover was commanded to drive its left-side wheels over a 10-cm rock (see \fig{fig:obstacle_negotiation_field_test}). Given the dominating performance of elastic suspensions (i.e., IE and MHS) during obstacle negotiation observed in our simulations, we now wanted to observe the potential differences in performance when dependency was introduced. Tests were performed at three different speeds (0.2, 0.5, and 1.0\,m/s) and each test was conducted three times for each speed and both the IE and MHS configurations. The magnitude of the force applied to the front wheels was recorded by an in-wheel force/torque sensor and vertical accelerations were measured by the same IMU as in the dynamic stability tests.

Table~\ref{table:test_acc} also gathers all the IMU measurements recorded during the obstacle negotiation tests. Both the IE and MHS configurations successfully overcome the obstacle at 0.2 and 0.5\,m/s but it was only with the MHS that the rover was capable of seamlessly negotiating the rock at 1.0\,m/s. At this speed, the observable impact on the IE configuration was such that no successful runs were ultimately conducted with this configuration due to concerns over safety and the structural integrity of the rover. 

\begin{figure}[!hbt]
     \begin{subfigure}[b]{0.49\linewidth}
         \centering
         \includegraphics[width=\linewidth]{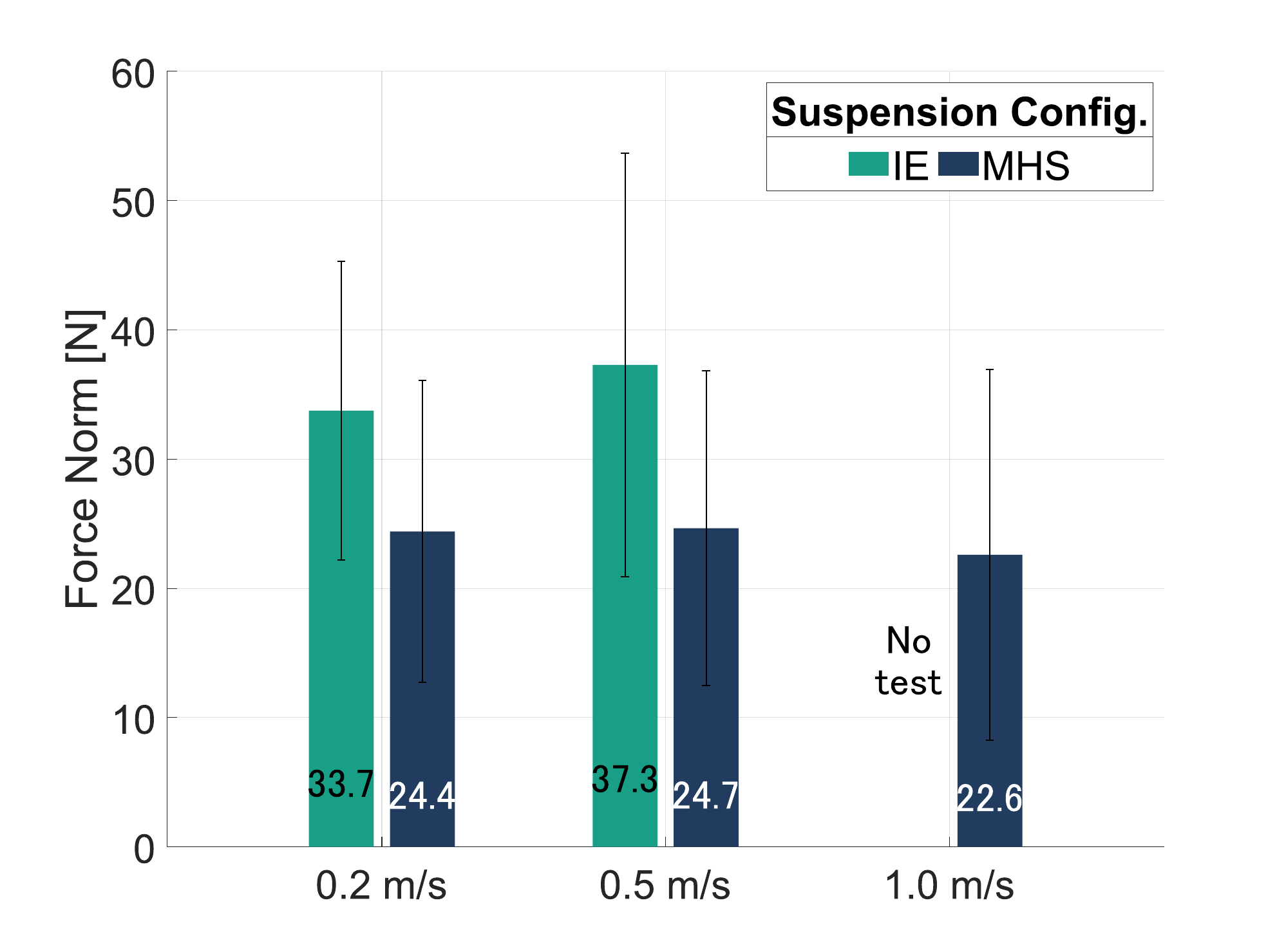}  
         \subcaption{Left front wheel.}\label{fig:left_wheel_force}
     \end{subfigure}
     \hfill
     \begin{subfigure}[b]{0.49\linewidth}
         \centering
         \includegraphics[width=\linewidth]{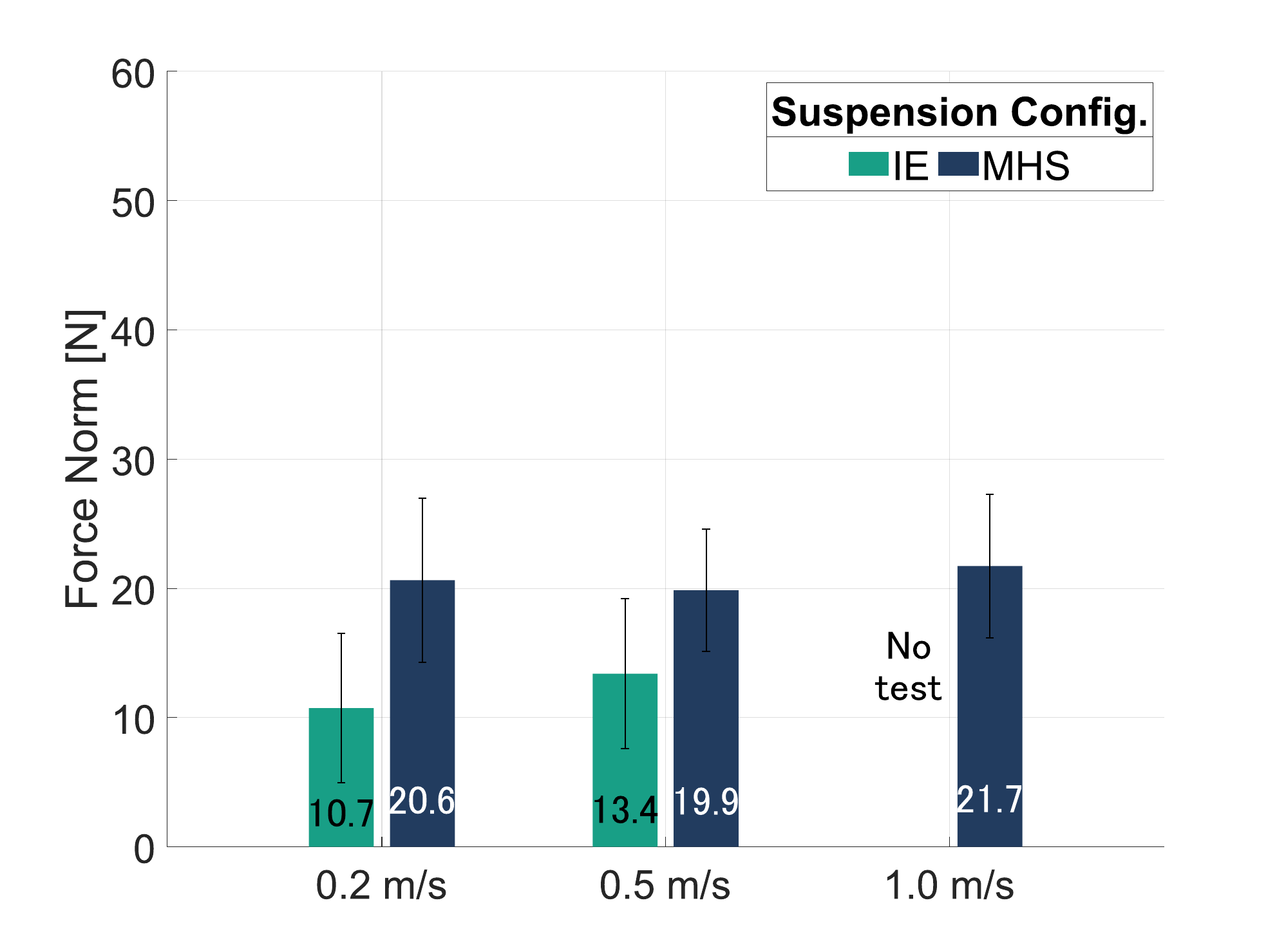}  
         \subcaption{Right front wheel.}\label{fig:right_wheel_force}
     \end{subfigure}
     \caption{Average of the norm of the force vector acting on both (a) left wheels (wheels overcoming the obstacle) and (b) right front wheels (opposite side wheels) when driving over a 10\,cm rock at different speeds with IE and MHS. Error bars indicate one standard deviation across all runs.}
     \label{fig:force_norm_comparison_in_obstacle_negotiation}
\end{figure}

\fig{fig:force_norm_comparison_in_obstacle_negotiation} display the average of the norm of the force vector acting on the front wheels when overcoming the rock for both the IE and MHS configurations. Dampening of impact loads on the front-left wheel (both mean and maximum force) was also greater in the case of the MHS and the degree of dampening increased with speed, reducing the loads by 24\% on average across wheels and speeds. The main benefit associated with the addition of dependency via the free-balancing rocker is the increased pressure exerted on the wheels not overcoming the obstacle, enabling the right side wheels to greater traction levels and drastically improving obstacle-surmounting capabilities of the rover (see \subfig{fig:force_norm_comparison_in_obstacle_negotiation}{fig:right_wheel_force}).

\subsection{Comparison with simulations}
Although field test conditions slightly diverge from the simulations (i.e., different gravity field and simplified wheel-terrain model), the results help us verify the suitability of our preliminary analysis. In the \textit{dynamic stability} scenario, similar trends are observed. The DR configuration experienced the most ineffective attenuation of vibrations and impact loads while the MHS  outperforms the other two. The degree of attenuation is, however, generally lower in real-world experiments than in the simulations, which can be attributed to the assumption of the ground surface as a non-deformable rigid body. In the field tests, part of the energy is dissipated in the compression and displacement of the sand beneath the wheels, an effect that was not modeled in our simulations. In the \textit{obstacle negotiation} case, the rate of reduction of the maximum impact loads ($-$8--48\% at 0.2--0.5\,m/s during testing) is within the range of those experienced during the simulation ($-$38\% at 1\,m/s). In this case, the contribution of the terrain model is negligible and, therefore, results have a higher resemblance between simulations and field testing.

\section{CONCLUSION}
\label{sec:conclusion}

The increased autonomy demanded by upcoming missions to the Moon and Mars implies planetary robots have to be capable of coping with a wide range of disturbances. The addition of compliant elements to the suspension system of these robots appeared to be vital in counteracting the detrimental effects of impact loads and vibrations when driving at high velocities ($\geq$\,1\,m/s) under weaker gravity fields. But even when these elements are included, the specific configuration of the suspension design plays an important role in the rover ultimate performance. A new passive suspension configuration, so-called mechanically-hybrid suspension (MHS), was proposed and compared with more traditional rocker and independent swing arm suspensions. The MHS combines the functional benefits of both dependent and elastic elements. Simulation results under a lunar-like gravity field confirmed our initial hypothesis. Field test results validated that an MHS configuration could greatly improve stability while successfully isolating the chassis of the rover from unwanted vibrations and impact loads beyond what could be accomplished with either of the other two commonly used passive configurations. An improved suspension design also affects other aspects involved in navigation, lowering the demand on perception and control systems, increasing duty cycles, and enabling higher levels of autonomy. Future work will explore additional improvements and variations in the suspension configuration such as combining the MHS with non-pneumatic, flexible wheels. Their combination could bring about higher levels of stability and terrain compliance while further reducing non-vertical impact loads and vibrations.

\section*{Acknowledgment}
The authors would like to thank Alan Allart, Tristan Lecocq, Kazuki Nakagoshi, Ryusuke Wada, Danishi Ai, and Merlijn Siffels for their invaluable help and support in the development of EX1.  

\bibliography{./IEEEabrv,bibliography.bib}

\begin{thebibliography}{10}
\providecommand{\url}[1]{#1}
\csname url@rmstyle\endcsname
\providecommand{\newblock}{\relax}
\providecommand{\bibinfo}[2]{#2}
\providecommand\BIBentrySTDinterwordspacing{\spaceskip=0pt\relax}
\providecommand\BIBentryALTinterwordstretchfactor{4}
\providecommand\BIBentryALTinterwordspacing{\spaceskip=\fontdimen2\font plus
\BIBentryALTinterwordstretchfactor\fontdimen3\font minus \fontdimen4\font\relax}
\providecommand\BIBforeignlanguage[2]{{%
\expandafter\ifx\csname l@#1\endcsname\relax
\typeout{** WARNING: IEEEtran.bst: No hyphenation pattern has been}%
\typeout{** loaded for the language `#1'. Using the pattern for}%
\typeout{** the default language instead.}%
\else
\language=\csname l@#1\endcsname
\fi
#2}}

\bibitem{Colaprete2010}
A.~Colaprete, P.~Schultz, J.~Heldmann, D.~Wooden, M.~Shirley, K.~Ennico, B.~Hermalyn, W.~Marshall, A.~Ricco, R.~C. Elphic, D.~Goldstein, D.~Summy, G.~D. Bart, E.~Asphaug, D.~Korycansky, D.~Landis, and L.~Sollitt, ``{Detection of water in the LCROSS ejecta plume},'' \emph{Science}, vol. 330, no. 6003, pp. 463--468, 2010.

\bibitem{Potts2015}
N.~J. Potts, A.~L. Gullikson, N.~M. Curran, J.~K. Dhaliwal, M.~K. Leader, R.~N. Rege, K.~K. Klaus, and D.~A. Kring, ``{Robotic traverse and sample return strategies for a lunar farside mission to the Schr{\"{o}}dinger basin},'' \emph{Advances in Space Research}, vol.~55, no.~4, pp. 1241--1254, 2015.

\bibitem{Bickler1989}
D.~B. Bickler, ``{Articulated Suspension Systems},'' 1989, {US} Pattent Office, Washington, D.C., US4840394.

\bibitem{Eisen1997}
H.~J. Eisen, C.~W. Buck, G.~R. Gillis-smith, and J.~W. Umland, ``{Mechanical design of the Mars Pathfinder Mission},'' in \emph{7th European Symposium}, Noordwijk, The Netherlands, 1997.

\bibitem{Gonzalez2018}
R.~Gonzalez and K.~Iagnemma, ``{Slippage estimation and compensation for planetary exploration rovers. State of the art and future challenges},'' \emph{Journal of Field Robotics}, vol.~35, no.~4, pp. 564--577, 2018.

\bibitem{Miller2002}
D.~P. Miller and T.~L. Lee, ``{High-speed traversal of rough terrain using a Rocker-Bogie mobility system},'' in \emph{Space 2002 and Robotics 2002}, Albuquerque, NM, 2002.

\bibitem{Wang2016}
S.~Wang and Y.~Li, ``{Dynamic Rocker-Bogie: Kinematical Analysis in a High-Speed Traversal Stability Enhancement},'' \emph{International Journal of Aerospace Engineering}, vol. 2016, pp. 1--8, sep 2016.

\bibitem{Iagnemma2003}
K.~Iagnemma, A.~Rzepniewski, S.~Dubowsky, and P.~Schenker, ``{Control of robotic vehicles with actively articulated suspensions in rough terrain},'' \emph{Autonomous Robots}, vol.~14, no.~1, pp. 5--16, 2003.

\bibitem{Bartlett2008}
P.~W. Bartlett, D.~Wettergreen, and W.~Whittaker, ``{Design of the Scarab Rover for Mobility {\&} Drilling in the Lunar Cold Traps},'' in \emph{International Symposium on Artificial Intelligence, Robotics and Automation in Space (i-SAIRAS)}, Hollywood, USA, 2008, pp. 3--6.

\bibitem{Cordes2014}
F.~Cordes, C.~Oekermann, A.~Babu, D.~Kuehn, T.~Stark, and F.~Kirchner, ``{An Active Suspension System for a Planetary Rover},'' in \emph{International Symposium on Artificial Intellifence, Robotics and Automation in Space (i-SAIRAS)}, Montreal, Canada, 2014.

\bibitem{Malenkov2016a}
M.~Malenkov, ``{Self-propelled automatic chassis of Lunokhod-1: History of creation in episodes},'' \emph{Frontiers of Mechanical Engineering}, vol.~11, no.~1, pp. 60--86, 2016.

\bibitem{NASA1973a}
NASA, ``{Apollo 17 Mission Report},'' Lyndon B. Johnson Space Center, Houston, Texas, Tech. Rep., 1973.

\bibitem{NASA1971a}
NASA, ``{Apollo 15 Lunar Roving Vehicle Systems Handbook},'' NASA, Houston, Texas, Tech. Rep., 1971.

\bibitem{RodriguezMartinez2019}
D.~Rodr{\'{i}}guez-Mart{\'{i}}nez, M.~{Van Winnendael}, and K.~Yoshida, ``{High-speed mobility on planetary surfaces: A technical review},'' \emph{Journal of Field Robotics}, vol.~36, no.~8, pp. 1436--1455, December 2019.

\bibitem{vrep2013}
E.~Rohmer, S.~P.~N. Singh, and M.~Freese, ``V-rep: A versatile and scalable robot simulation framework,'' in \emph{IEEE/RSJ International Conference on Intelligent Robots and Systems}, 2013, pp. 1321--1326.

\bibitem{Gonzalez2019a}
R.~Gonzalez, D.~Apostolopoulos, and K.~Iagnemma, ``{Improving rover mobility through traction control: simulating rovers on the Moon},'' \emph{Autonomous Robots}, vol.~43, no.~8, pp. 1977--1988, 2019.

\bibitem{RodriguezMartinez2019a}
D.~Rodr{\'i}guez-Mart{\'i}nez, F.~Buse, M.~{Van Winnendael}, and K.~Yoshida, ``{The effects of increasing velocity on the tractive performance of planetary rovers},'' in \emph{ISTVS 15th European-African Regional Conference}, Prague, Czech Republic, 2019.

\bibitem{Biesiadecki2006a}
J.~Biesiadecki, E.~Baumgartner, R.~Bonitz, B.~Cooper, F.~Hartman, P.~Leger, M.~Maimone, S.~Maxwell, A.~Trebi-Ollennu, E.~Tunstel, and J.~Wright, ``{Mars exploration rover surface operations: driving opportunity at Meridiani Planum},'' \emph{IEEE Robotics {\&} Automation Magazine}, vol.~13, no.~2, pp. 63--71, 2006.

\bibitem{orr2022effects}
S.~Orr, J.~Casler, J.~Rhoades, and P.~de~Le{\'o}n, ``Effects of walking, running, and skipping under simulated reduced gravity using the {NASA} {Active} {Response} {Gravity} {Offload} {System} ({ARGOS}),'' \emph{Acta Astronautica}, vol. 197, pp. 115--125, 2022.

\end{thebibliography}

\end{document}